%% file: main.tex
\documentclass[10pt,twocolumn,letterpaper]{article}

\usepackage[pagenumbers]{iccv} 

\usepackage{multirow}
\usepackage{pifont}
\usepackage{subcaption}


\usepackage{xcolor}
\definecolor{darkgreen}{RGB}{0,100,0}
\definecolor{mediumgreen}{RGB}{0,128,0}
\definecolor{lightgreen}{RGB}{144,238,144}
\definecolor{limegreen}{RGB}{50,205,50}
\definecolor{forestgreen}{RGB}{34,139,34}

\definecolor{iccvblue}{rgb}{0.21,0.49,0.74}
\usepackage[pagebackref,breaklinks,colorlinks,allcolors=iccvblue]{hyperref}

\def\paperID{8478} 
\def\confName{ICCV}
\def\confYear{2025}

\title{StreamGS: Online Generalizable Gaussian Splatting Reconstruction \\ for Unposed Image Streams}

\newcommand{\mysep}{\hspace{0.5cm}}
\author{Yang Li\textsuperscript{1,*} \mysep Jinglu Wang\textsuperscript{1}\mysep Lei Chu\textsuperscript{1}\mysep Xiao Li\textsuperscript{1}\mysep Shiu-Hong Kao\textsuperscript{1,2,*}  \mysep Ying-Cong Chen\textsuperscript{2,3} \mysep Yan Lu\textsuperscript{1} \\
\textsuperscript{1}Media Computing Group, Microsoft Research Asia \\
\textsuperscript{2}CSE Dept., HKUST \mysep \textsuperscript{3}AI Thrust, HKUST(GZ) \\
{\tt\small yangliaftermath@gmail.com} \mysep {\tt\small skao@cse.ust.hk} \mysep {\tt\small yingcongchen@ust.hk} \\
{\tt\small \{jinglu.wang, leichu, li.xiao, yanlu\}@microsoft.com}
}
\begin{document}
\maketitle
\renewcommand{\thefootnote}{}
\footnotetext{\textsuperscript{*}Work done during the internship at MSRA.}
\renewcommand{\thefootnote}{\arabic{footnote}}
\input{sec/0_abstract}    
\input{sec/1_intro_mod}

\input{sec/1-2-relatedwork}
\input{sec/2_method_mod}
\input{sec/3_exp}

{
    \small
    \bibliographystyle{ieeenat_fullname}
    \bibliography{main}
}



\end{document}

%% file: sec/0_abstract.tex
\begin{abstract}

The advent of 3D Gaussian Splatting (3DGS) has advanced 3D scene reconstruction and novel view synthesis. With the growing interest of interactive applications that need immediate feedback, online 3DGS reconstruction in real-time is in high demand. However, none of existing methods yet meet the demand due to three main challenges: the absence of predetermined camera parameters, the need for generalizable 3DGS optimization, and the necessity of reducing redundancy. We propose \textbf{StreamGS}, an online generalizable 3DGS reconstruction method for unposed image streams, which progressively transform image streams to 3D Gaussian streams by predicting and aggregating per-frame Gaussians. Our method overcomes the limitation of the initial point reconstruction \cite{dust3r} in tackling out-of-domain (OOD) issues by introducing a content adaptive refinement. 
The refinement enhances cross-frame consistency by establishing reliable pixel correspondences between adjacent frames. Such correspondences further aid in merging redundant Gaussians through cross-frame feature aggregation. The density of Gaussians is thereby reduced, empowering online reconstruction by significantly lowering computational and memory costs. Extensive experiments on diverse datasets have demonstrated that StreamGS achieves quality on par with optimization-based approaches but does so 150 times faster, and exhibits superior generalizability in handling OOD scenes.

\end{abstract}

%% file: sec/1_intro_mod.tex
\section{Introduction}
\label{sec:intro}


\begin{figure}
    \centering
    \includegraphics[width=0.8\linewidth]{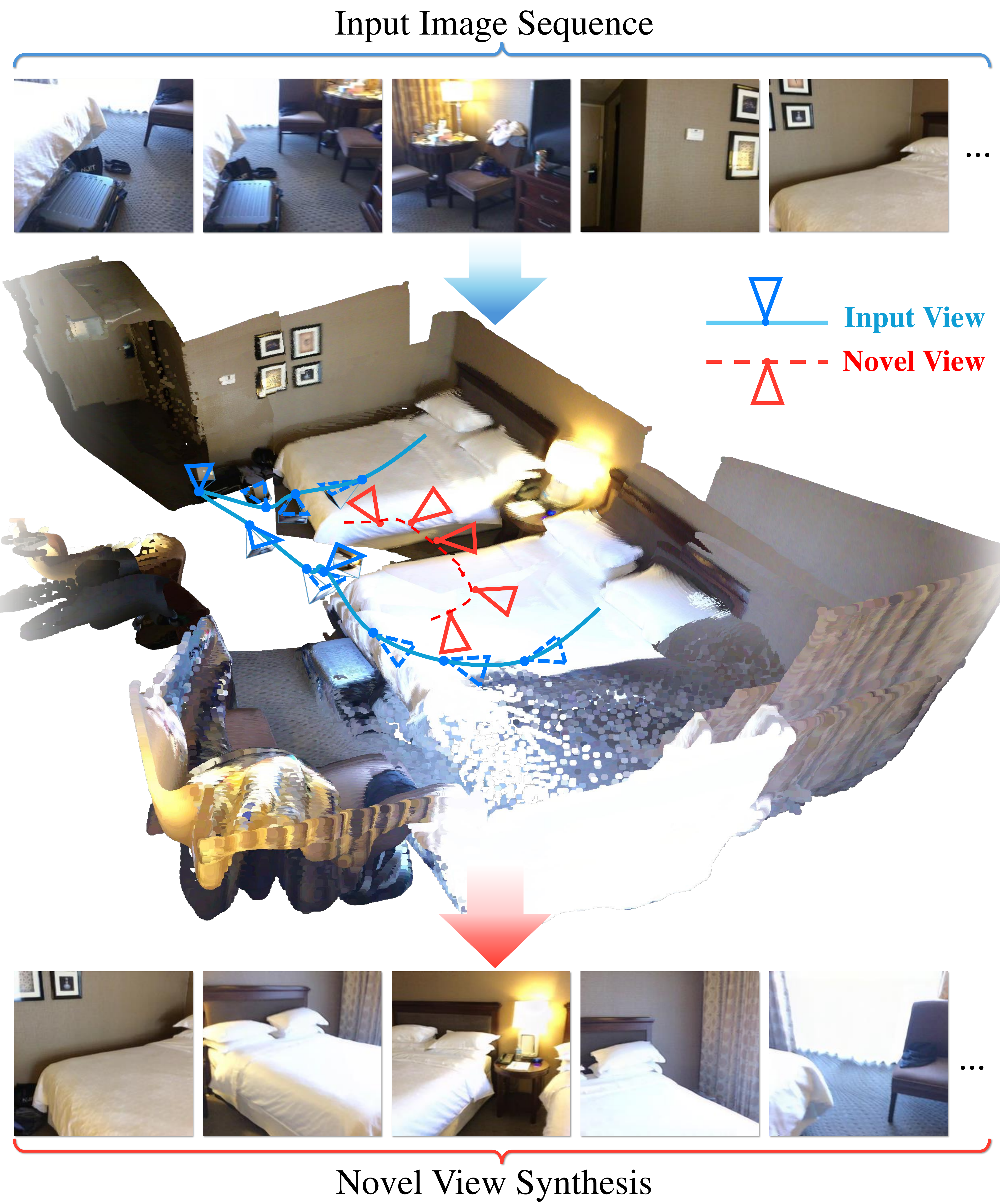}
    \caption{The proposed StreamGS efficiently transforms image streams into Gaussian streams by progressively reconstructing and aggregating per-frame 3D Gaussians. We show our reconstructed 3DGS (visualized as points) alongside estimated camera poses (in blue), and synthesized novel views.}
    \label{fig:teaser}
\end{figure}

The field of 3D Scene reconstruction \cite{surfelnerf, yan2024gsslam} for novel view synthesis from image streams has gained increasing attention, due to its significance in enabling interactive applications that offer users instant feedback. In this context,  the advent of 3D Gaussian Splatting (3DGS) \cite{3dgs} marks a major advancement in high-quality, real-time rendering.  This progress highlights the urgency for efficient, on-the-fly generation of 3DGS from image streams. 

Nonetheless, online 3DGS reconstruction faces unique challenges.
(1) \textbf{Unknown camera.}	The conventional preprocessing using Structure from Motion (SfM) \cite{schonberger2016structure} for camera estimation is impractical for real-time streaming. This is due to the absence of the full image set and the time-consuming computation. Thus, there is a need for methods that can operate without pre-determined cameras.
(2) \textbf{Generalizability.}	3DGS reconstruction requires multi-iteration optimization, impractical for online applications due to the need for all images in advance. This restricts the development of a generalizable method that can process image streams in a feed-forward manner.
(3) \textbf{Redundancy.}	The significant overlap between frames leads to high redundancy when reconstructing Gaussians individually for each frame, increasing the streaming process's resource intensity.

Scene reconstruction without known cameras has been explored in SLAM-based \cite{zhang2024improved, zou2022mononeuralfusion, zhu2024nicer, yan2024gsslam, yugay2023gaussianslam} and NeRF-based \cite{ bian2023nopenerf, scnerf} methods. Only a few 3DGS-based methods are discussed \cite{cfgs, park2024splinegs}. These methods primarily consider camera poses as learnable parameters and optimized alongside Gaussians through the iterative optimization. Yet, this optimization-driven approach for aligning each frame considerably increases the reconstruction time, rendering it impractical in the online scenario.

Recently, generalizable 3DGS reconstruction has been investigated for sparse views \cite{splatter-image,pixelsplat,mvsplat}.
These approaches transform pixels into Gaussians, whose parameters are decoded from images via 2D encoder-decoder networks. They achieve the 3DGS reconstruction for each image in a single feed-forward pass. However, these generalizable approaches are primarily designed for monocular or binocular settings, suited to sparse-view inputs. In addition, multi-view methods \cite{pixelsplat, mvsplat} typically infer Gaussian centers through stereo matching, highly dependent on known cameras, limiting their applicability for online scenarios with numerous unposed images. Furthermore, they combine multi-view 3DGS sets by simply uniting them, which overlooks cross-view alignment and overlaps, leading to misalignment and redundancy issues within image streams. The recent method \cite{instantsplat} performs Gaussian downsampling in 3D space to reduce redundancy, but it necessitates traversing and processing Gaussians in 3D grids, which is also time-consuming and is highly dependent on the grid resolution.

Emerging models like DUSt3R \cite{dust3r} and MASt3R \cite{mast3r} enable sparse-view geometry reconstruction by simultaneously predicting 3D points and estimating camera parameters. These advances pave the way for more efficient, feed-forward, pose-free 3D reconstruction methods.
A straightforward approach to generalize 3DGS reconstruction is to add a Gaussian predictor to the DUSt3R-like framework. However, this introduces several challenges. These models usually require datasets with ground truth 3D geometry, which may not always be available. Additionally, applying pretrained models to out-of-domain (OOD) data can result in inaccurate pose and 3D point estimations. Moreover, generating 3D points for each frame individually causes redundancy due to overlapping adjacent frames, potentially leading to ghosting artifacts from pose estimation errors.

In this paper, we introduce \textbf{StreamGS}, a novel pipeline for online, generalizable 3DGS reconstruction from unposed image streams. StreamGS aims to progressively construct and update the 3DGS representation of the scene frame-by-frame, in a feed-forward manner, as illustrated in \cref{fig:teaser}. We leverage the pretrained DUSt3R to initially predict 3D point for the current frame using the previous frame as a reference. However, this initialization may encounter inaccuracies due to OOD issues. To mitigate this, we capitalize the insight that adjacent frames offer sufficient correspondences to refine the reconstruction. Unlike DUSt3R using predicted 3D points to establish correspondences, we adopt content-adaptive descriptors for more reliable matching, allowing for the \textbf{adaptive refinement} of the reconstruction by enhancing consistency between adjacent views. Furthermore, such correspondences help to prune redundant pixel-aligned Gaussians. Correlated pixel-wise features across frames are effectively aggregated, removing duplicates and achieving \textbf{adaptive density control}. Finally, we decode Gaussians from such aggregated features. StreamGS is adept at predicting and integrating Gaussians for the current frame into the existing Gaussian set seamlessly with a feed-forward pass.
In summary, our contribution is three-fold.
\begin{itemize} 
\item We introduce a novel pipeline for the online, generalizable reconstruction of image streams without requiring camera parameters, marking a first in this field.
\item The proposed adaptive refinement enhance cross-frame consistency of 3DGS reconstruction, and the adaptive density control mechanism minimizes adjacent-view redundancy, thereby highly reducing computational costs in online reconstruction.
\item Upon evaluation across diverse datasets, our method achieves high novel view synthesis quality comparable to the optimization-based method \cite{cfgs} but with 150x faster reconstruction speed. Additionally, our method outperforms existing pose-dependent generalizable 3DGS methods in handling OOD scenes, showing superior generalizability. 

\end{itemize}

%% file: sec/1-2-relatedwork.tex
\section{Related Works}
\paragraph{Generalizable 3D Gaussian Splatting.} Many recent studies aim to propose generalizable 3D-GS methods capable of predicting Gaussians within a single feed-forward pass. These works can be classified into two main categories: single-view reconstruction and multi-view reconstruction. Single-view reconstruction does not involve pose estimation as there are no multi-view constraints. Inspired by the insight from LRM \cite{lrm} that large transformer-based \cite{vaswani2017attention} backbone networks can learn 3D priors from large-scale 3D data, the potential of predicting Gaussians from a single image in a single feed-forward pass has been comprehensively explored. Numerous feed-forward models have been proposed, such as GRM \cite{grm}, TriplaneGS \cite{triplanegs}, and GMamba \cite{shen2024gamba}. However, these methods are primarily not applicable to multi-view scenarios as they always assume canonical poses.

Concurrently, many works focusing on multi-view inputs follow a similar paradigm, such as GS-LRM \cite{zhang2025gslrm}, LGM \cite{lgm}, and MVGMamba \cite{yi2024mvgamba}. They concatenate the input images with camera embeddings like Pl\"{u}cker rays to facilitate the network in learning the proper fusion of Gaussians from multi-view inputs. However, these large 3D backbone networks mostly perform well only on synthetic objects due to the shortage of large-scale scene-level 3D data in the real world. Referring to generalizable NeRF methods like pixelNeRF \cite{yu2021pixelnerf}, other research turns to multi-view stereo (MVS) matching to locate or initialize the centers of Gaussians, with other attributes decoded using a lightweight 2D encoder. Representative works with this design include MVSGaussian \cite{liu2025mvsgaussian}, pixelSplat \cite{pixelsplat}, and MVSplat \cite{mvsplat}. However, both camera ray embedding in large transformer-based models and stereo matching rely on known poses and intrinsics of each input view. Another main limitation of these methods is that they focus on sparse-view inputs. PixelSplat \cite{pixelsplat} and MVSplat \cite{mvsplat} only support up to three views. Therefore, existing generalizable 3D-GS models cannot address the problem of feed-forward reconstruction from endless image streams.

\paragraph{Pose-free 3D Gaussian Splatting.} Recently, many works have aimed to eliminate the need for Structure-from-Motion (SfM) preprocessing steps using COLMAP \cite{schonberger2016structure} software. Following the design of previous pose-free NeRF methods like NoPe-NeRF \cite{bian2023nopenerf}, Lu-NeRF \cite{lunerf}, and localRF \cite{localrf}, CF-3DGS \cite{cfgs} introduces depth priors into the optimization of 3D-GS and performs progressive reconstruction. As each new image arrives, CF-3DGS optimizes both the pose and 3D Gaussians of the input image based on its depths, aiming to align the 3DGS from the new image with the preceding reconstruction. However, it still relies on known camera intrinsics. CF-3DGS is also not robust, as the accuracy of depth priors significantly impacts its reconstruction quality, limiting its application to common scenes. Moreover, it depends on thousands of optimization steps for each view, significantly extending the reconstruction time for each scene. Compared to generalizable 3D-GS models, current pose-free methods are so inefficient that they cannot be applied to image streams with a large number of frames.

\paragraph{Online 3D reconstruction of image streams.} Online 3D reconstruction has been extensively studied in the field of SLAM \cite{zhu2022nice, zhang2024improved, zou2022mononeuralfusion, zhu2024nicer, yan2024gsslam, yugay2023gaussianslam}. However, these methods typically involve additional information and most leverage SDF and NeRF as scene representations. NICE-SLAM \cite{zhu2022nice} uses RGB-D streams as input, with the reconstructed scene represented by NeRF. NICER-SLAM \cite{zhu2024nicer} relies on geometric priors, including surface normals and depths. SurfelNeRF \cite{surfelnerf} focuses on the novel view synthesis quality of online reconstruction with RGB streams, but it requires the poses and intrinsics of frames. Gaussian-SLAM \cite{yugay2023gaussianslam} reconstructs the scene using 3D-GS, but it also relies on RGB-D streams.

%% file: sec/2_method_mod.tex
\section{Methods}
\begin{figure*}[h]
    \centering
    \includegraphics[width=\linewidth]{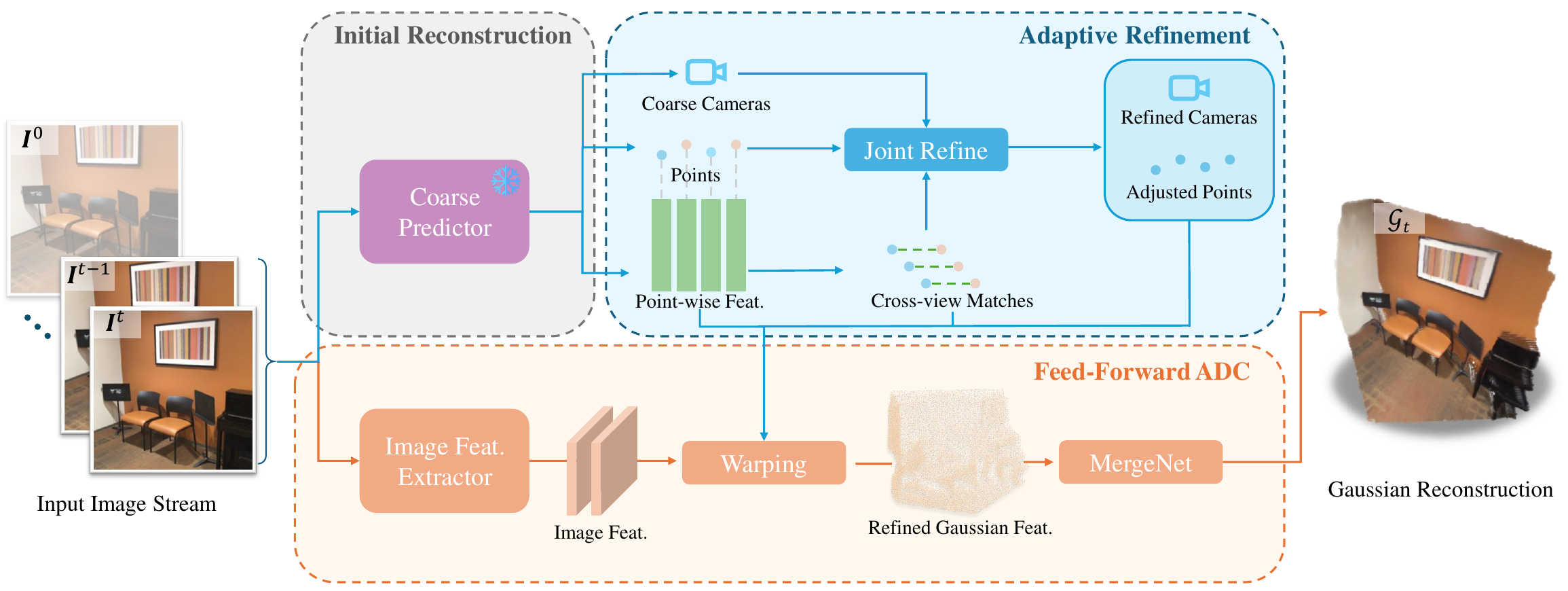}
    \caption{\textbf{Method overview.} Our StreamGS progressively reconstruct and aggregate 3D Gaussians from the unposed image stream. Given the adjacent image pair $(\mathbf{I}^{t-1}, \mathbf{I}^{t})$, we first perform the initial reconstruction that predicts pixel-wise 3D points with their features and coarse camera poses, using a pretrained coarse predictor. 
    Since the coarse predictions may suffer from OOD issues, we refine both the camera poses and 3D positions by establishing new point-wise correspondences.
    We aggregate cross-frame image and 3D features by warping and merging to reduce redundancy. Finally we decode the aggregated features to Gaussian primitives.
    }
    \vspace{-15pt}
    \label{fig:pipeline}
\end{figure*}

Given a sequence of \textit{unposed} images over time, our objective is to progressively reconstruct the 3D Gaussian Splatting (3DGS) representation in an online manner. Specifically, at each timestamp $t$, the goal is to derive the 3DGS $\mathcal{G}^t$, which encapsulates the 3D scene aggregated from the images $\mathcal{I}^t = \{\mathbf{I}^i\}_{i=1}^{t}$.

~\cref{fig:pipeline} illustrates our overall framework. In order to make reconstruction efficient with limited computational resources, we employ an incremental construction strategy. At each time step $t$, we focus on generating the 3DGS $\mathbf{G}^t$ of the current frame and merge it with the previous accumulated reconstruction $\mathcal{G}^{t-1}$ to obtain the full reconstruction $\mathcal{G}^{t}$ of the current time step. Specifically, with current frame $\mathbf{I}^t$, we use $\mathbf{I}^{t-1}$ as reference frame and estimate the point maps and cameras of each using an \textit{initial reconstruction} module. With newly established matches between current and reference views, we further refine the quality of the points and the cameras in the \textit{adaptive refinement} module. Finally, we generate $\mathbf{G}^t$ using the refined points and features of $\mathbf{I}^t$, and merge it with previous reconstructions according to the established matches, achieving the \textit{feed-forward adaptive density control (ADC)}.

\subsection{Preliminaries}
\label{sec:pre}

\paragraph{Gaussian splatting.}

Previous work \cite{3dgs} represents a scene or object using a set of Gaussian distributions. Specifically each gaussian primitive could be denoted as $G(x; \mu, \Sigma)=e^{-\frac{1}{2}(x-\mu)^T\Sigma^{-1}(x-\mu)}$ and the covariance $\Sigma$ is decomposed into the rotation matrix $R$ and scaling matrix $S$ to ensure the positive semi-definiteness during optimization, that is $\Sigma = RSS^TR^T$. The view-dependent color of the appearance is represented by a set of spherical harmonics (SH) coeffients and oppacity value $\alpha$.

\subsection{Initial Two-view Reconstruction}
\label{sec:initial}

Given the current frame $\mathbf{I}^t$ and a reference frame $\mathbf{I}^{t'}$, we use a coarse predictor $\phi_{3D}$ and estimate the point map $\mathbf{X}^{t|t'}$ of the current frame under the local coordinate system of the reference frame together with its corresponding confidence map $\mathbf{C}^{t|t'}$, where the superscript $\cdot|t'$ indicates that the local coordinate system adapts to that of $\mathbf{I}^{t'}$. Intuitively, the predicted point map stores the 3D point coordinate that the pixel unprojected to in, and the confidence maps measure the certainty of point maps at each pixel, reflecting a prior-based estimation of reconstruction accuracy and difficulty. Formally, we define

\begin{equation}
    (\mathbf{X}^{t|t'},\mathbf{X}^{t'|t'}, \mathbf{C}^{t|t'} ,\mathbf{C}^{t'|t'}) = \phi_{3D}(\mathbf{I}^{t}, \mathbf{I}^{t'}).
\end{equation}

In order to help align the current local-coordinate pointmap with the global coordinate, we also get the output in the local coordinate system of $\mathbf{I}^{t}$ by computing $\phi_{3D}(\mathbf{I}^{t'}, \mathbf{I}^{t})$.  With these predicted point maps, we could further estimate the camera matrix $\mathbf{P}=\mathbf{K}[\mathbf{R}|\mathbf{t}]$, composed of its intrinsic parameters $\mathbf{K}$, and its extrinsic parameters, the rotation matrix $\mathbf{R}$ and the translation vector $\mathbf{t}$. Specifically we assume that the principle points $\mathbf{c}$ are centered and pixels are squares and have
\begin{equation}
\hat{f^t} =  \arg \min\limits_{f} \Sigma_{\mathbf{p}\in\mathbf{I}^t}  \parallel \mathbf{p} - \mathbf{c} - f \dfrac{(\mathbf{x}_{\mathbf{p}}, \mathbf{y}_{\mathbf{p}})}{\mathbf{z}_{\mathbf{p}}} \parallel,
\label{eq:focal-raw}
\end{equation}
where $(\mathbf{x}_{\mathbf{p}}, \mathbf{y}_{\mathbf{p}}, \mathbf{z}_{\mathbf{p}})\in \mathbf{X}^{t|t}$ is the 3D point coordinate that the pixel $\mathbf{p}$ unprojected to. Moreover, we approximate relative pose $\mathbf{P}^{t} = [\mathbf{R}, \mathbf{t}]$ of $\mathbf{I}^{t}$ to $\mathbf{I}^{t'}$ by solving the following points registration problem:
\begin{equation}
\resizebox{.9\hsize}{!}{$
\begin{aligned}
 [\mathbf{R}^{t} \vert \mathbf{t}^{t}] &= \arg \min\limits_{s, \mathbf{R}, \mathbf{t}} \sum_{\mathbf{p} \in \mathbf{I}^
 {t-1}} \mathbf{C}^t(\mathbf{p}) \parallel s(\mathbf{R} \mathbf{X}^{t'|t'}(\mathbf{p}) + \mathbf{t}) - \mathbf{X}^{t'|t}(\mathbf{p}) \parallel^2, \\
     &\, \mathbf{C}^t = \mathbf{C}^{t'|t'} \odot\mathbf{C}^{t'|t}, 
\end{aligned}
$}
\label{eq:pts-reg}
\end{equation}
where $\odot$ is the Hadamard product and $s$ is the scale factor.

In our implementation, we leverage DUSt3R \cite{dust3r} as the coarse predictor due to its effective pretraining on dedicated 3D scene datasets and its efficiency in reconstructing 3D points. For each time step $t$, we process the pair $(\mathbf{I}^{t-1}, \mathbf{I}^{t})$ to derive the 3D points $ (\mathbf{X}^{t|t-1},\mathbf{X}^{t-1|t-1})$ corresponding to both frames referenced in the coordinate frame $t-1$, the reversed pair $(\mathbf{I}^{t}, \mathbf{I}^{t-1})$ to obtain the same points $(\mathbf{X}^{t-1|t},\mathbf{X}^{t|t})$ but in the coordinate frame of $t$, which will be reused at the next timestamp to boost efficiency. For simplicity, and without loss of generality, the following discussion will focus on the reconstruction of the image pair $(\mathbf{I}^{t-1}, \mathbf{I}^{t})$. Note that training $\phi_{3D}$ requires 3D geometric supervision, which may not be available in our monocular video input scenario. Thus, we leverage the pretrained $\phi_{3D}$ from \cite{dust3r}.
\subsection{Adaptive Refinement}
\label{sec:refine}
We note that the initial reconstruction quality is compromised due to the OOD challenge as the coarse predictor is frozen. This observation motivates us to enhance reconstruction through content adaptation, with the goal of adaptively refining both poses and 3D reconstructions.

We perform the adaptive refinement based on establishing new robust matches between adjacent frames. We employ a matching head, $\phi_{match}$, to extract local 3D features, denoted as $(\mathbf{F}^{t-1}_{3D}, \mathbf{F}^{t}_{3D}) \in \mathbb{R}^{H \times W \times d}$, from the consecutive image pair $(\mathbf{I}^{t-1}, \mathbf{I}^{t})$.
The matches between the two images can be established through nearest reciprocal (NN) searching, satisfying the following condition:
\begin{equation}\label{eq:match}
\resizebox{.9\hsize}{!}{$
\begin{aligned}
 \mathcal{M}^{t-1,t} &= \{ i_k \leftrightarrow j_k \vert i_k = \mathrm{NN} (j_k) \ \mathrm{and}\ j_k = \mathrm{NN}(i_k) \}_{k=1}^{N}, \\
    & \mathrm{s.t.}\ \mathrm{NN}(i_k) = \underset{0 \leq j_k \leq H \times W}{\arg \min} \mid 1 - \cos <\mathbf{F}^{t-1}_{3D, i_k}, \mathbf{F}^{t}_{3D, j_k}> \mid,
\end{aligned}
$}
\end{equation}
where $i_k, j_k$ are pixel index in image $\mathbf{I}^{t-1}, \mathbf{I}^{t}$ respectively, and the measurement of feature distance is cosine similarity. With the correspondences found, a residual transform $\boldsymbol{\Delta} = [\Delta\mathbf{R}, \Delta\mathbf{t}]$ can be re-estimated following \cref{eq:pts-reg} by only taking matched points into account. Then we apply the residual transform to the pointmap $\mathbf{X}^{t,t-1}$ to retain refined $\tilde{\mathbf{X}}^{t|t-1}$, and the pose of $\mathbf{I}^{t}$ is updated to $\tilde{\mathbf{P}}^{t}$ by performing PnP-RANSAC \cite{pnp, ransac} on 3D-2D correspondences derived from matches.

\paragraph{Gaussian decoding.} 
We directly predict the other parameters of 3D Gaussians at each pixel with a light-weight decoder $\phi_{GS}$ as:
\begin{equation}\label{eq:gs_decoder}
\begin{aligned}
    \mathbf{G}^i = [\mathbf{q}^i, \boldsymbol{s}^i, \alpha^i, \mathbf{c}^i] = \phi_{GS}(\mathbf{F}_{gs}), \\
    \mathbf{F}^i_{GS} = \mathbf{F}^i_{2D} \oplus \mathbf{X}^i \oplus \mathbf{F}^i_{3D}, \quad \mathbf{F}_{2D} = \phi_{2D}(\mathbf{I}^i), 
\end{aligned}
\end{equation}
where $i = \{t-1,t\}$, $\oplus$ denotes the channel-wise concatenation, $\phi_{2D}$ denotes the 2D image feature extractor, $\mathbf{q}^i \in \mathbb{R}^{H \times W \times 4}$ and $\mathbf{s}^i \in \mathbb{R}^{H \times W \times 3}$ represent rotation quaternions and scales of pixel-aligned Gaussians at image $\mathbf{I}^i$. 
We incorporate an additional image feature extractor because the coarse predictor is frozen and cannot be trained by our monocular video setting. Extracting new image features is essential for decoding Gaussians, especially for texture-related properties. Experiments show the importance of the image feature extractor.
Then covariance matrix is built with $\boldsymbol{\Sigma}^i = \mathbf{R}(\mathbf{q}^i)\mathbf{s}\mathbf{s}^{\mathrm{T}} \mathbf{R}(\mathbf{q}^i)^\mathrm{T}$. 
Note that $\bar{\mathbf{G}}$ is not the final Gaussians since it needs to be merged into the previous Gaussian set following \cref{sec:adc}.

\subsection{Feed-Forward ADC}
\label{sec:adc}
With pixel-aligned Gaussian parameters $\mathbf{G}^t$ of $T$ images, previous methods  \cite{pixelsplat, mvsplat, splatter-image, lgm} always naively take the union of Gaussians in all images as the final prediction, i.e., $\mathcal{G} = \bigcup_{t=1}^T \mathbf{G}^i$ and  $\vert \mathcal{G} \vert  = T \times H \times W$. However, this approach is both memory-intensive and inefficient in rendering, particularly when dealing with the continuous input of video frames during online reconstruction. Our key observation is that the  matched Gaussian pairs in neighboring frames are excessive and prunable since they consistently share similar attributes in shape and color, and are closely distributed, which is validated in \cref{tab:similarity}. 
It is noted that we have already acquired dense pixel-wise matches between neighboring frames from \cref{eq:match}. Therefore, we propose a novel feed-forward Adaptive Density Control  strategy based on revisiting these dense correspondences.

\paragraph{Feature aggregation.} The primary advantage of pixel-based correspondences is that they are able to convert the computationally intensive 3D Gaussian aggregation process into a more efficient 2D pixel-wise one. This significantly enhances computational efficiency. Initially, the feature of Gaussian parameters $\mathbf{F}_{gs}^{t}$ of the frame $\mathbf{I}^t$ can be aligned to the previous frame $\mathbf{I}^{t-1}$ using the following wrapping:
\begin{equation}
    \mathbf{F}_{GS}^{t|t-1}(j) = \begin{cases} 
        \mathbf{F}_{GS}^{t}(k) & \text{if } (j, k) \in \overline{\mathcal{M}}^{t-1,t} \\
        \mathbf{F}_{GS}^{t-1}(j) & \text{else,} \\
        \end{cases}
\end{equation}
where $\mathbf{F}_{GS}^{t|t-1}$ denotes the feature of Gaussian primitives in the frame $\mathbf{I}^{t}$ aligned to $\mathbf{I}^{t-1}$. Instead of using the raw correspondences set $\mathcal{M}^{t-1,t}$ in \cref{eq:match}, we use the extended set $\overline{\mathcal{M}}^{t-1,t}$, which includes matches of neighboring pixels such as $(i+1, j+1)$ in addition to the initial $(i,j)$. It serves as a type of anti-aliasing technique to reduce void pixels within the wrapped feature map. As for the unmatched pixels, we simply replicate the corresponding feature vector within $\mathbf{F}_{gs}^{t-1}$. With the aligned feature, the Gaussian feature of two frames can be merged by modifying \cref{eq:gs_decoder}, taking the form:
\begin{equation}
    \hat{\mathbf{G}}^{t|t-1} = [\hat{\mathbf{q}}^{t}, \hat{\boldsymbol{s}}^{t}, \hat{\alpha}^{t}, \hat{\mathbf{c}}^{t}] = \phi_{MG}(\mathbf{F}_{GS}^{t|t-1} \oplus \mathbf{F}_{GS}^{t-1}),
\end{equation}
where $\phi_{MG}$ denotes the MergeNet that simply consists of two convolutional layers, which merges features and decodes them to Gaussian primitives.
In this way, every matched pair of Gaussians between $\mathbf{I}^{t}$ and $\mathbf{I}^{t-1}$ is aggregated into a single one, highly reducing the number of Gaussians. 
The final aggregated Gaussian set is $\mathcal{G}^t = \mathcal{G}^{t-1} \cup \mathbf{G}^{t|t-1}$.

Without gradient back-propagation of rendering loss in the original 3DGS paper \cite{3dgs}, our ADC process runs exceptional fast. In terms of the input group, the final prediction of Gaussian primitives $\mathcal{G}^{t}$ consists of the merged Gaussians and the unmatched Gaussians at each frame.

\subsection{Loss Functions}
\label{sec:opt}
The optimization of the 2D feature extractor $\phi_{2D}$, the Gaussian decoder network, $\phi_{GS}$ involves both a rendering loss function and a reconstruction loss function:
\begin{equation}\label{eq:opt}
\resizebox{.9\hsize}{!}{$
    \begin{aligned} \mathcal{L}(\mathbf{I}^i, \hat{\mathbf{I}}^i, \hat{\mathbf{I}}^i_M) &= \mathcal{L}_{\mathrm{render}}(\mathbf{I}^i, \hat{\mathbf{I}}^i) + \mathcal{L}_{\mathrm{recon}}(\hat{\mathbf{I}}^i, \hat{\mathbf{I}}^i_M) \\ &= \parallel \mathbf{I}^i - \hat{\mathbf{I}}^i \parallel_2 + \lambda \parallel \mathbf{I}^i - \hat{\mathbf{I}}^i \parallel_{\mathrm{LPIPS}} + \parallel \hat{\mathbf{I}}_{\mathrm{M}}^i - \hat{\mathbf{I}}^i \parallel_2, \end{aligned},
    $}
\end{equation}
where  $\mathbf{I}^i$ is the ground truth frame, $\hat{\mathbf{I}}^i$ is the rendered image with full Gaussian primitives before the merge process, and $\hat{\mathbf{I}}_{\mathrm{M}}^i$ is the rendered image with merged Gaussians. Due to the lightweight nature of the two networks, the algorithm converges quickly within thousands of steps. The reconstruction loss term facilitates the merge network in fusing the pixel-aligned Gaussian parameters from different frames, aiming to render the same frame as before the merge process but with significantly fewer Gaussian primitives.


%% file: sec/3_exp.tex
\section{Experiments}
StreamGS operates on an image stream of a scene, jointly predicting the corresponding poses and Gaussians in a feed-forward manner. To assess its performance, we evaluate our method on the task of novel view synthesis from monocular videos, as detailed in Sec. \ref{sec:exp-nvs}. Additionally, we validate the effectiveness of the proposed alignment module and the efficiency of the Gaussians merge process, as described in Sec. \ref{sec:exp-abla}.
\vspace{-15pt}
\paragraph{Baselines and datasets.} To the best of our knowledge, StreamGS is the first method to reconstruct unposed videos in a feed-forward manner. Consequently, we compare our method separately with pose-free 3DGS works and generalizable splatting methods, including pixelSplat \cite{pixelsplat}, MVSplat \cite{mvsplat}, and CF-3DGS \cite{cfgs}. PixelSplat and MVSplat are representative methods of generalizable Gaussian Splatting, but they both rely on known camera poses and intrinsics. In contrast, CF-3DGS is a pose-free method but requires optimization loops to align poses and Gaussians. For a comprehensive comparison, we evaluate the methods on large-scale datasets with diverse scenes, including RE10K \cite{re10k}, ACID \cite{acid}, ScanNet \cite{scannet}, DL3DV \cite{dl3dv}, and MVImgNet \cite{mvimgnet}. Each dataset comprises monocular video sequences with per-frame camera pose annotations. 
\vspace{-15pt}
\paragraph{Implementation details.} Both the 2D image feature extractor $\phi_{2D}$ and the MergeNet $\phi_{GS}$ are double-layer convolutional networks. For fairness, StreamGS and other generalizable methods are trained on the identical training split of the RE10K dataset using the Adam \cite{adam} optimizer with the same learning rate of $2 \times 10^{-4}$ and a cosine scheduler. The parameter $\lambda$ in Eq. \ref{eq:opt} is set to $0.05$. All methods are trained for 30K iterations and tested on a single NVIDIA Tesla A100 80GB GPU. The images are resized to $224 \times 224$ and the batch size is set to 14. For the non-generalizable method CF-3DGS, we follow its original setting \cite{cfgs}. Note that CF-3DGS is evaluated only on a subset of the full test set due to its low reconstruction efficiency, as shown in \cref{fig:time_efficiency}. Unlike pose-dependent methods, CF-3DGS and our method require poses of novel views for rendering. CF-3DGS freezes the trained Gaussian model and performs additional optimization steps to learn the poses. Our method, as described in \cref{sec:refine}, carries out pose alignment and refinement processes to estimate the poses. More details can be referred to supplementary materials.

\begin{table}
\centering
    \resizebox{0.45\textwidth}{!}{
    \begin{tabular}{@{}lccccc}
        \toprule
        & mean $\downarrow$ & Opa. $\downarrow$& Rot. $\downarrow$ & Scale $\downarrow$ & SH $\downarrow$ \\ \midrule
        Random Pick & 0.41 & 0.10 & 3.31 & 2.47 & 0.27 \\
        \textbf{Matched Pairs} & \textbf{0.18} &\textbf{ 0.04} & \textbf{3.06} & \textbf{1.28} & \textbf{0.04} \\
        \bottomrule
    \end{tabular}}
        \caption{Similarities of attributes between matched GS across adjacent frames on RE10K. Random Pick refers to \textbf{randomly} picking two GS from two frames separately, while Matched Pairs refers to our \textbf{matched} GS defined in \cref{eq:match}. }\label{tab:similarity}
        \vspace{-15pt}
\end{table}

\subsection{Novel View Synthesis}
\label{sec:exp-nvs}
\subsubsection{Reconstruction Quality}

\begin{table*}[t]
\centering
\setlength{\tabcolsep}{3pt} 
\renewcommand{\arraystretch}{1.2} 
\resizebox{\textwidth}{!}{
\begin{tabular}{@{}lcccccccccccccccccc@{}}
\toprule
 & \multirow{3}{*}{\textbf{PF}} & \multirow{3}{*}{\textbf{G}} & \multicolumn{3}{c}{RE10K \cite{re10k} (Source Domain)} & \multicolumn{3}{c}{DL3DV \cite{dl3dv}} & \multicolumn{3}{c}{MVImgNet \cite{mvimgnet}} & \multicolumn{3}{c}{ScanNet \cite{scannet}} & \multicolumn{3}{c}{ACID \cite{acid}} \\ \cmidrule(lr){4-6} \cmidrule(lr){7-9} \cmidrule(lr){10-12} \cmidrule(lr){13-15} \cmidrule(lr){16-18}
& & &PSNR $\uparrow$ & LPIPS $\downarrow$ & SSIM $\uparrow$ & PSNR $\uparrow$ & LPIPS $\downarrow$ & SSIM $\uparrow$ & PSNR $\uparrow$ & LPIPS $\downarrow$ & SSIM $\uparrow$ & PSNR $\uparrow$ & LPIPS $\downarrow$ & SSIM $\uparrow$ & PSNR $\uparrow$ & LPIPS $\downarrow$ & SSIM $\uparrow$\\ \midrule
pixelSplat \cite{pixelsplat} & \ding{56} & \ding{52} & 22.64 & 0.21 & 0.75 & 19.43 & 0.39 & 0.50 & 19.22 & 0.49 & 0.60 & 27.56 & 0.19 & 0.82 & \textbf{29.93} & \textbf{0.14} & \textbf{0.84}\\
MVSplat \cite{mvsplat}       & \ding{56}& \ding{52} & \textbf{23.54} & \textbf{0.13}& \textbf{0.90} & 17.84 & 0.36 & 0.45 & 16.29 & 0.50 & 0.53 & 26.25 & 0.19 & 0.82 & 28.83 & 0.14 & 0.86\\
CF-GS \cite{cfgs}               & \ding{52} & \ding{56} & 23.46 & 0.24 & 0.75 & 19.93 & 0.31 & 0.62 & \textbf{26.33} & \textbf{0.33} & \textbf{0.88} & 22.88 & 0.42 & 0.76 & 28.16 & 0.19 & 0.81\\
\textbf{StreamGS} & \ding{52} & \ding{52} & 22.42 & 0.17 & 0.83 & \textbf{20.54} & \textbf{0.24} & \textbf{0.64} & 25.05 & 0.31 & 0.79 & \textbf{28.43} & \textbf{0.16} & \textbf{0.86} & 28.50 & 0.15 & 0.84\\
\bottomrule
\end{tabular}}
\caption{Quantitative comparison with existing state-of-art methods on Novel View Synthesis of monocular videos. \textbf{PF} indicates whether the method is \textbf{pose-free}. \textbf{G} indicates whether the methods is generalizable. Our method consistently achieves scores comparable to state-of-the-art methods that are either not pose-free or lack generalizability. In the table, the best result is highlighted in \textbf{bold}.}
\label{tab:comparison}
\vspace{-12pt}
\end{table*}

\paragraph{Quantitative comparison.} We compare StreamGS with baseline methods on the quality and efficiency of novel view synthesis. Following previous 3D-GS research \cite{3dgs}, we report PSNR, SSIM \cite{ssim}, and LPIPS \cite{lpips} as metrics of reconstruction quality. The quantitative results are shown in \cref{tab:comparison}. On the source domain RE10K \cite{re10k}, existing state-of-the-art generalizable methods demonstrate competitive scores, with MVSplat even outperforming the optimization-based CF-3DGS. However, their performance degrades significantly on out-of-domain datasets such as DL3DV \cite{dl3dv} and MVImgNet \cite{mvimgnet}, as these datasets contain various scene types, including outdoor environments and more complex indoor scenes with different objects and illumination conditions compared to RE10K. Naturally, CF-3DGS, which is based on thousands of optimization steps, performs well on the aforementioned datasets. However, its PSNR slightly decreases on ScanNet due to more irregular camera movements and increased motion blur in the frames, posing a challenge for CF-3DGS in recovering camera poses. PixelSplat performs well on ScanNet, thanks to its robust feature extraction backbone trained on ImageNet \cite{imagenet}, while MVSplat achieves the lowest score. According to the table, StreamGS consistently achieves scores comparable to all other methods. Its PSNR on MVImgNet and DL3DV is significantly higher than that of PixelSplat and MVSplat, even without given poses and intrinsics, demonstrating our method's superior generalizability on unseen datasets with significant domain gaps.

\begin{figure*}
    \centering
    \noindent\makebox[\textwidth]{%
        \includegraphics[width=1\textwidth]{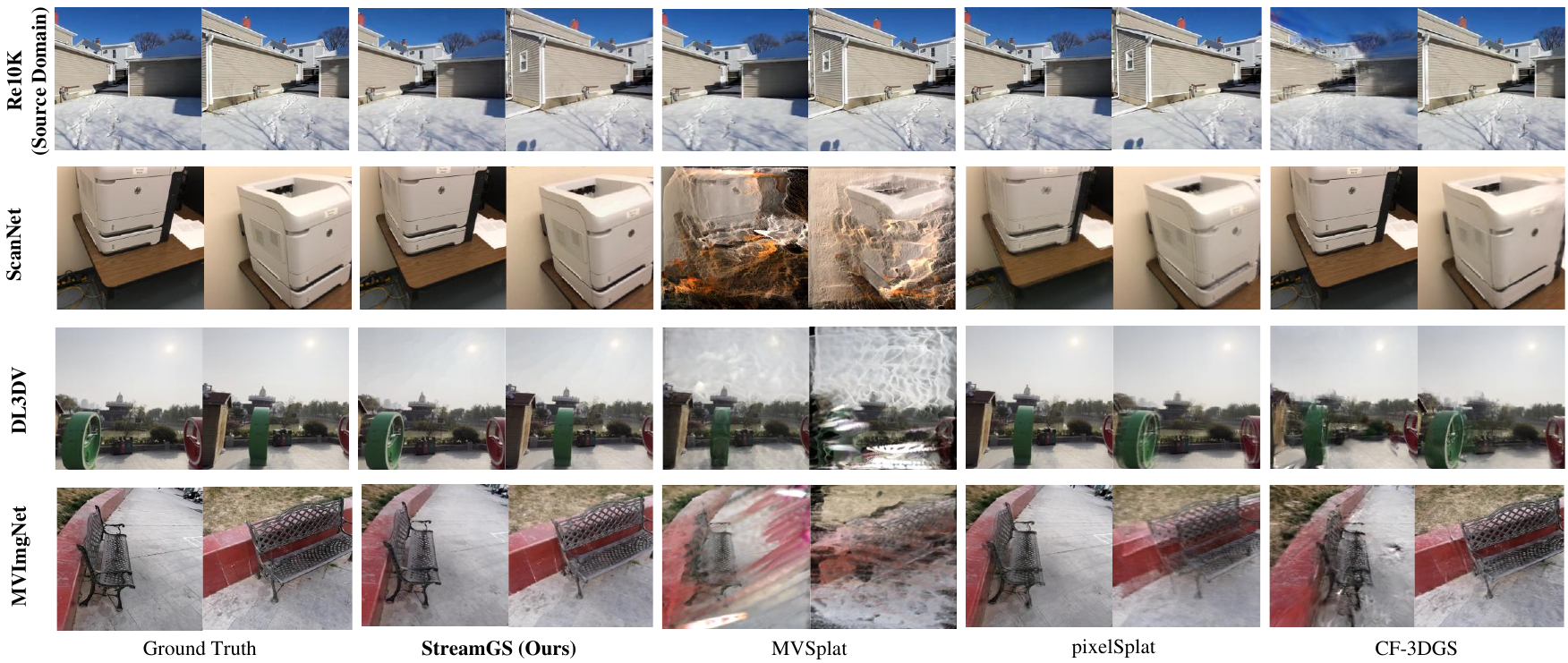}
    }
    \vspace{-15pt}
    \caption{Qualitative comparison on novel view synthesis. We show the results on both source domain, RE10K \cite{re10k}, and other domains, ScanNet \cite{scannet}, DL3DV \cite{dl3dv} and MVImgNet \cite{mvimgnet}. All generalizable methods are trained only on RE10K and tested on the other datasets. StreamGS outperforms other methods in several challenging scenarios, especially for the out-of-domain data. }\label{fig:visual_comp}
    \vspace{-15pt}
\end{figure*}

\begin{figure*}[ht]
    \centering
        \noindent\makebox[\textwidth]{%
        \includegraphics[width=1\textwidth]{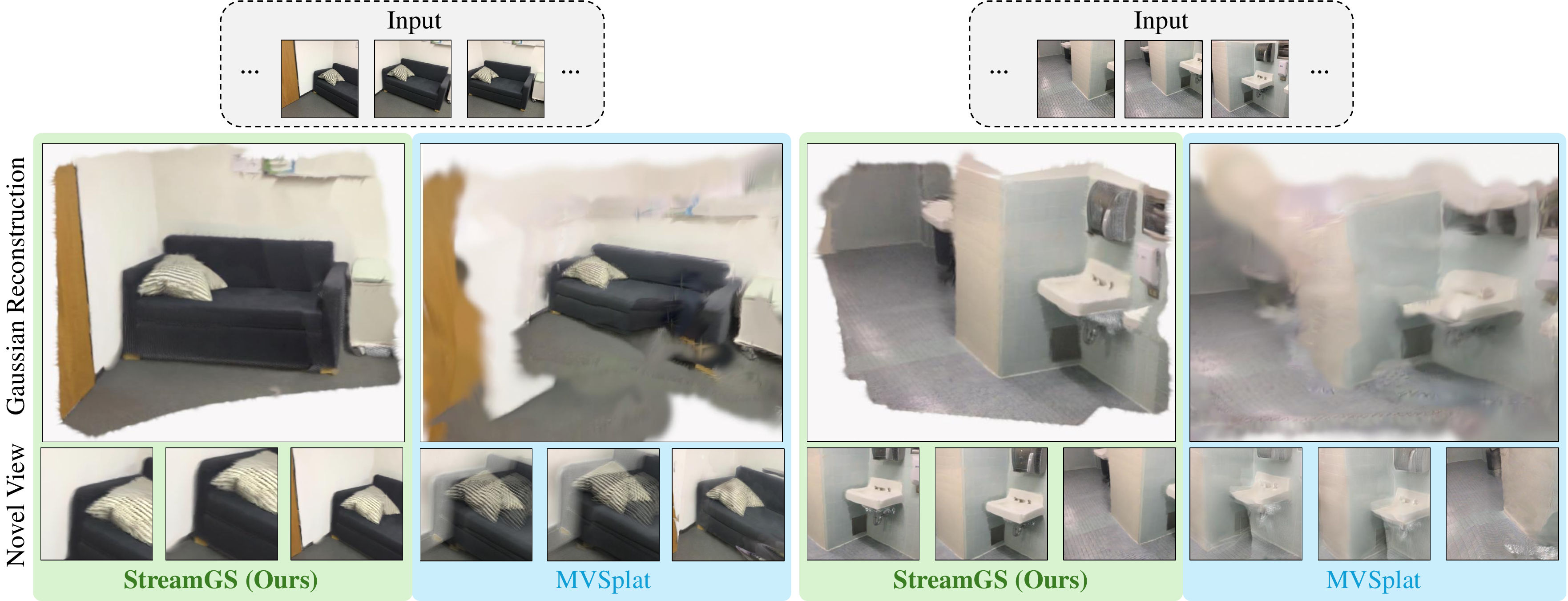}
    }
    \caption{Visual comparison of Gaussian reconstruction and novel view synthesis from image streams with ScanNet \cite{scannet} dataset. Unlike MVSplat \cite{mvsplat}, which struggles with view aggregation, our results show significantly better visual quality on OOD data. Note that Our StreamGS and MVSplat are both trained with RE10K \cite{re10k} data, and MVSplat needs predetermined cameras.}
    \label{fig:vis}
    \vspace{-1.5em}
\end{figure*}
\vspace{-10pt}
\paragraph{Qualitative comparison.} A qualitative comparison with state-of-the-art methods is also presented in \cref{fig:visual_comp} and \cref{fig:vis}. While all methods demonstrate high-quality novel view rendering on RE10K \cite{re10k}, our method exhibits superior robustness on out-of-domain datasets. Due to the significant domain gap including texture, illumination and camera motion, both MVSplat and pixelSplat fail to predict accurate depths for the printer shown in the second row, resulting in severe floating artifacts in the rendered image. The third row shows the reconstruction of a plaza from DL3DV \cite{dl3dv}. Similarly, since outdoor scenes are much less represented in RE10K \cite{re10k}, MVSplat struggles to extract stereo cues from text-less skies and objects with disparate illumination, deteriorating the rendering quality. The final row shows a bench scene in MVImgNet, on which baseline generalizable methods also fails due to the similar domain gap issues. These cases also demonstrate that the generalizability of pixelSplat surpasses that of MVSplat. As the figure shows, CF-3DGS also does not perform well on some outdoor scenes. In contrast, the visual quality of our method on out-of-domain datasets remains high.

\subsubsection{Reconstruction Efficiency}

\begin{figure}[!htp]
    \centering
    \includegraphics[width=\linewidth]{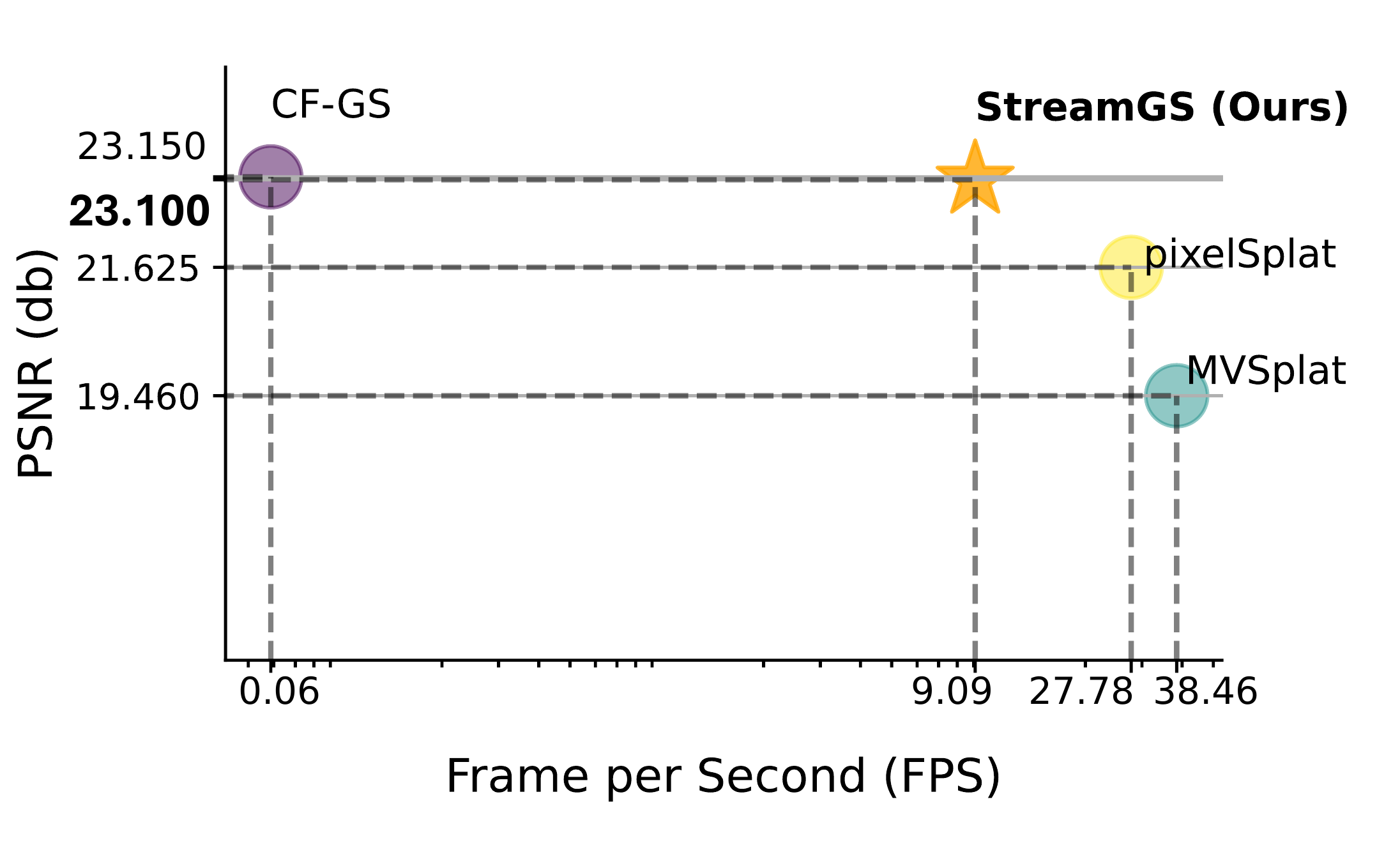}
  \vspace{-3em}
  \caption{Reconstruction speed measured by frames processed per second (FPS). The x-axis is log-scaled for the better visualization. }
  \label{fig:time_efficiency}
  \vspace{-20pt}
\end{figure}

\begin{table}[htbp]
\centering
\setlength{\tabcolsep}{3pt} 
\renewcommand{\arraystretch}{1.2} 
\resizebox{0.96\linewidth}{!}{
\begin{tabular}{@{}lcccccccc@{}}
\toprule
 & \multicolumn{2}{c}{Coarse Pred (\cref{sec:initial})} & \multicolumn{2}{c}{Ada. Refine (\cref{sec:refine})} & \multicolumn{2}{c}{ADC (\cref{sec:adc})} & \multicolumn{2}{c}{Total} \\ \cmidrule(lr){2-3} \cmidrule(lr){4-5} \cmidrule(lr){6-7} \cmidrule(lr){8-9} 
& time/s & param/M & time/s & param/M & time/s & param/M & time/s & param/M \\ \midrule
StreamGS & 0.02 & 656.74 & 0.08 & 1.83 & 0.01 & 0.04 & 0.11 & 658.61 \\
\bottomrule
\end{tabular}}
\caption{Efficiency metrics of each component.}\label{tab:runtime}
\vspace{-30pt}
\end{table}

In addition to rendering quality, we also compare the reconstruction efficiency of our method with baseline models. \cref{fig:time_efficiency} illustrates a plot of reconstruction quality, measured by the average PSNR reported in \cref{tab:comparison}, versus efficiency, measured by the processing time per frame (FPS). The figure shows that StreamGS achieves second place with a PSNR of 23.1, only 0.05 lower than CF-3DGS \cite{cfgs}. This indicates that our method achieves nearly the same rendering quality as the best model. However, thanks to the feed-forward design, StreamGS is \textbf{150 times} faster than CF-3DGS, predicting Gaussian primitives for up to 9 frames within one second. Without known camera information, our method involves additional alignment and pose estimation processes across frames, which limits the inference speed, making it slower than MVSplat \cite{mvsplat} and pixelSplat \cite{pixelsplat}. However, according to the scores reported in \cref{tab:comparison} and \cref{fig:time_efficiency}, our method is more generalizable and achieves better rendering quality than these methods. \cref{tab:runtime} shows the efficiency of each component of our method.

\subsection{Ablation Study}\label{sec:exp-abla}
In this section, we discuss the effectiveness of our main design about joint refinement (in Sec. \ref{sec:refine}) and feed-forward ADC module (in Sec \ref{sec:adc}). More ablation studies on framework design can be found in supplementary materials.
\subsubsection{Effectiveness of Joint Refinement}

\begin{table}[h]
\scalebox{0.9}{
\centering
\setlength{\tabcolsep}{3pt} 
\renewcommand{\arraystretch}{1.2} 
\resizebox{0.5\textwidth}{!}{
\begin{tabular}{@{}lcccccc@{}}
\toprule
 & \multicolumn{3}{c}{RE10K \cite{re10k}} & \multicolumn{3}{c}{ACID \cite{acid}} \\ \cmidrule(lr){2-4} \cmidrule(lr){5-7} 
& PSNR $\uparrow$ & LPIPS $\downarrow$ & SSIM $\uparrow$ & PSNR $\uparrow$ & LPIPS $\downarrow$ & SSIM $\uparrow$ \\ \midrule
w/o refine. & 17.25 & 0.32 & 0.58 & 17.23 & 0.39 & 0.48 \\
\textbf{w/ refine} & \textbf{22.42} &\textbf{ 0.17} & \textbf{0.83} & \textbf{28.50} & \textbf{0.15} & \textbf{0.84} \\
\bottomrule
\end{tabular}}
}
\vspace{-5pt}
\caption{Evaluation of effectiveness of joint refinement.}\label{tab:wo_refine}
\vspace{-12pt}
\end{table}

Joint refinement of cameras and centers of Gaussians plays a crucial role in the success of our method. To validate the effectiveness of joint refinement, we conduct an ablation study by skipping the refinement process during inference. In other words, the poses and intrinsics are directly estimated by Eq. \ref{eq:focal-raw} and \ref{eq:pts-reg}. \cref{tab:wo_refine} shows the quantitative comparison between the two settings. Without joint refinement, Gaussian primitives are cast from shifted origins of the camera with erroneous orientations, and the poses of novel views are also inaccurate, causing the rendered images to be shifted and distorted. This severely deteriorates the rendering quality, as shown in \cref{tab:wo_refine}. The PSNR of the images rendered with direct estimation decreases by 23.06\% and 38.62\% on the RE10K \cite{re10k} and ACID \cite{acid} datasets, respectively.

\subsubsection{Effectiveness of Gaussian Merging Process}
\begin{table}[h]
\scalebox{0.9}{
\centering
\setlength{\tabcolsep}{3pt} 
\renewcommand{\arraystretch}{1.2} 
\resizebox{0.5\textwidth}{!}{
\begin{tabular}{@{}lcccccc@{}}
\toprule
 & \multicolumn{2}{c}{MVImgNet \cite{mvimgnet}} & \multicolumn{2}{c}{ACID \cite{acid}} \\ \cmidrule(lr){2-3} \cmidrule(lr){4-5} 
& Compress. Ratio $\uparrow$ & PSNR $\uparrow$   &Compress. Ratio $\uparrow$ & PSNR $\uparrow$  \\ \midrule
w/o merge. & 1.00 & 25.70 & 1.00 & 29.21  \\
\textbf{merge all} & \textbf{1.58} & 25.05 &\textbf{ 1.68 }& 28.07  \\
\bottomrule
\end{tabular}}
}
\caption{Evaluation of the efficiency improvements of the Gaussian merging process and its impact on rendering quality.}\label{abla:wo_merge}
\vspace{-10pt}
\end{table}

During the feed-forward ADC, StreamGS prunes pixel-aligned Gaussians through a merging process. To evaluate the memory efficiency improvement and its impact on rendering quality, we compare the average number of Gaussians per frame and PSNR before and after the merging process. We define the compression ratio of Gaussians during the merging process as the ratio of the average number of Gaussians per frame, i.e., $H \times W$, to that after the merging process. The metrics are reported in \cref{abla:wo_merge}. The results demonstrate that the merging process can prune Gaussians per frame by 36.71\% and 40.48\% on MVImgNet \cite{mvimgnet} and ACID \cite{acid}, respectively. Meanwhile, the PSNR scores after the merging process only slightly decrease by 2.53\% and 3.90\%, respectively. This ablation study demonstrates that the designed Gaussian merging process efficiently reduces memory usage during reconstruction and rendering, with a negligible impact on reconstruction quality.

\section{Conclusion}
We propose a novel and holistic generalizable pose-free reconstruction pipeline named \textit{StreamGS}, dedicated to the online reconstruction of endless unposed image streams, such as monocular videos. To the best of our knowledge, our method is the first generalizable model capable of predicting Gaussians corresponding to the input stream in a feed-forward manner, without relying on known poses and intrinsics. Compared to pose-free but optimization-based methods, our method achieves comparable reconstruction quality while reducing the learning time to within several milliseconds, avoiding optimization steps. Compared to other generalizable methods, StreamGS eliminates the dependence on poses and intrinsics and manages to reconstruct more accurate scenes on out-of-domain datasets, demonstrating better domain generalizability.
\vspace{-7pt}
\paragraph{Limitations.} Although our method runs fast, joint refinement process still includes additional time costs, making it slower than MVSplat \cite{mvsplat}. Moreover, our approach encounters common reconstruction challenges, including texture-less regions and long sequences.